\documentclass[10pt,twocolumn,letterpaper]{article}

\usepackage{iccv}
\usepackage{times}
\usepackage{epsfig}
\usepackage{graphicx}
\usepackage{amsmath}
\usepackage{amssymb}

\usepackage{multirow}
\usepackage{xcolor}
\usepackage{enumitem}
\usepackage{bm}
\usepackage{color}
\usepackage{makecell}
\usepackage[pagebackref=true,breaklinks=true,letterpaper=true,colorlinks,bookmarks=false,citecolor=blue,linkcolor=blue]{hyperref}
\newcommand{\eat}[1]{}

\newcommand\blfootnote[1]{%
	\begingroup 
	\renewcommand\thefootnote{}\footnote{#1}%
	\addtocounter{footnote}{-1}%
	\endgroup 
}
\usepackage[breaklinks=true,bookmarks=false]{hyperref}

\iccvfinalcopy 

\def\iccvPaperID{2560} 

\ificcvfinal\fi

\begin{document}

\title{TransVG: End-to-End Visual Grounding with Transformers}

\author{Jiajun Deng$^\dagger$, Zhengyuan Yang$^\ddagger$, Tianlang Chen$^\ddagger$, Wengang Zhou$^{\dagger,\S}$, Houqiang Li$^{\dagger,\S}$\\
\small $^\dagger$ CAS Key Laboratory of GIPAS, University of Science and Technology of China, Hefei, China\\
\small $^\ddagger$ University of Rochester \\
\small $^\S$ Institute of Artificial Intelligence, Hefei Comprehensive National Science Center \\
{\tt\small dengjj@mail.ustc.edu.cn}
}

\maketitle
\ificcvfinal\thispagestyle{empty}\fi

\begin{abstract}
   In this paper, we present a neat yet effective transformer-based framework for visual grounding, namely TransVG, to address the task of grounding a language query to the corresponding region onto an image. The state-of-the-art methods, including two-stage or one-stage ones, rely on a complex module with manually-designed mechanisms to perform the query reasoning and multi-modal fusion. However, the involvement of certain mechanisms in fusion module design, such as query decomposition and image scene graph, makes the models easily overfit to datasets with specific scenarios, and limits the plenitudinous interaction between the visual-linguistic context. To avoid this caveat, we propose to establish the multi-modal correspondence by leveraging transformers, and empirically show that the complex fusion modules (\eg, modular attention network, dynamic graph, and multi-modal tree) can be replaced by a simple stack of transformer encoder layers with higher performance. Moreover, we re-formulate the visual grounding as a direct coordinates regression problem and avoid making predictions out of a set of candidates (\emph{i.e.}, region proposals or anchor boxes). Extensive experiments are conducted on five widely used datasets, and a series of state-of-the-art records are set by our TransVG. We build the benchmark of transformer-based visual grounding framework and make the code available at \url{https://github.com/djiajunustc/TransVG}.
\end{abstract}

\blfootnote{*Corresponding Author: Wengang Zhou and Houqiang Li.} 

\vspace{-0.1in}
\section{Introduction}	

\begin{figure}[t]
	\centering {\includegraphics[width=0.48\textwidth]{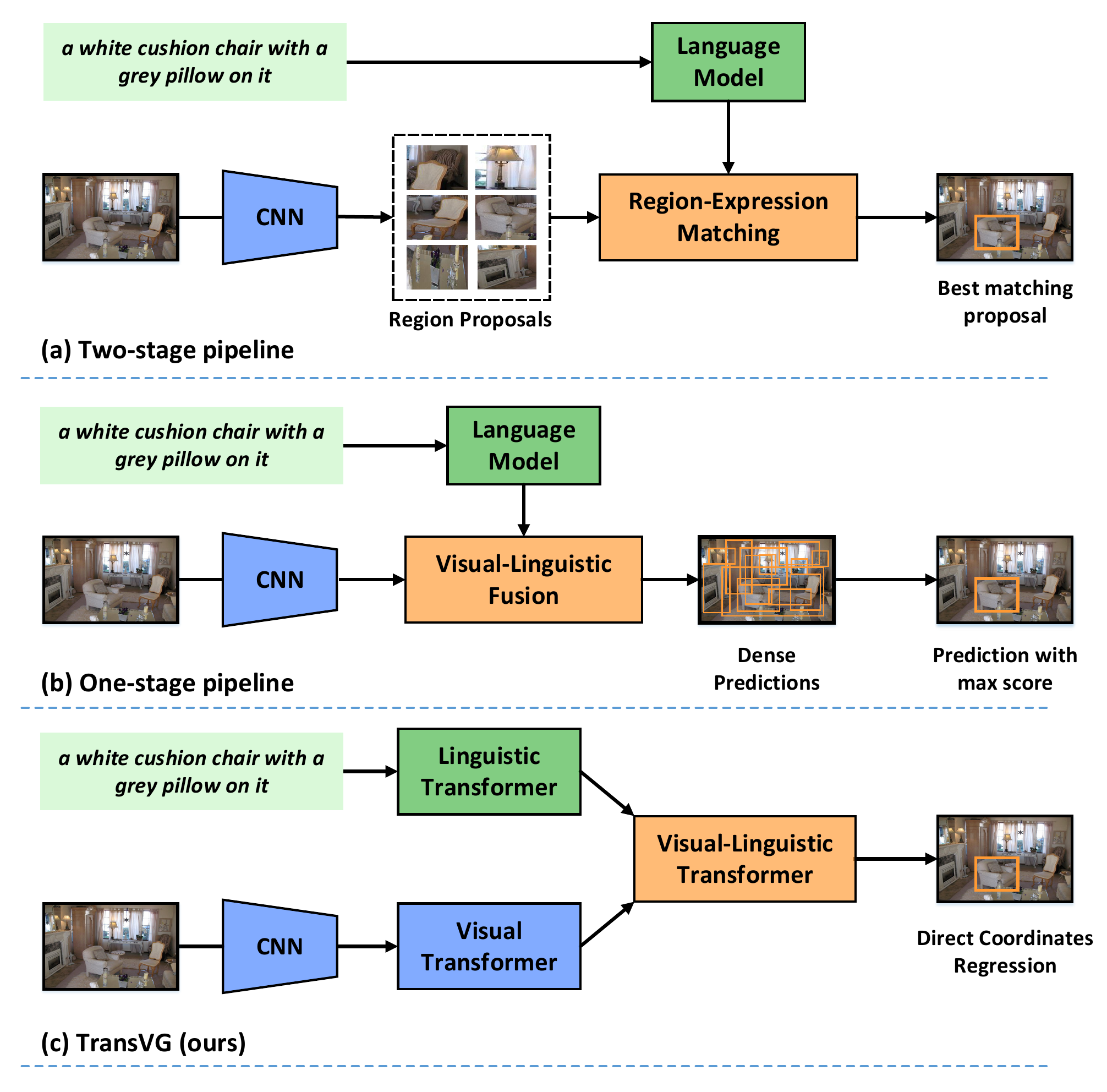}}
	
	\caption{A comparison of (a) two-stage pipeline, (b) one-stage pipeline, and (c) our proposed TransVG framework. TransVG performs intra-modality and inter-modality relation reasoning with a stack of transformer layers in a homogeneous way, and grounds the object by directly regressing the box coordinates.}	
	
	\label{fig:intro}
\end{figure}

Visual grounding (also known as referring expression comprehension~\cite{mao2016generation,yu2016modeling}, phrase localization~\cite{kazemzadeh2014referitgame,plummer2017flickr30k}, and natural language object retrieval~\cite{hu2016natural,li2017deep}) aims to predict the location of a region referred by the language expression onto an image. The evolution of this technique is of great potential to provide an intelligent interface for the natural language expression of human beings and the visual components of the physical world. Existing methods addressing this task can be broadly grouped into the two-stage and one-stage pipelines shown in 
Figure~\ref{fig:intro}. In specific, the two-stage approaches~\cite{mao2016generation,nagaraja2016modeling,wang2019learning,yu2016modeling} first generate a set of sparse region proposals and then exploit region-expression matching to find the best one. The one-stage approaches~\cite{chen2018real,liao2020real,yang2019fast} perform visual-linguistic fusion at intermediate layers of an object detector and output the box with the maximal score over pre-defined dense anchors.

Multi-modal fusion and reasoning is widely studied in the literature~\cite{antol2015vqa,pan2020x,wang2021multi,yang2021causal,zhang2020object}, and it is the core problem in visual grounding. In general, the early two-stage and one-stage methods address multi-modal fusion in a simple way. Concretely, the two-stage Similarity Net~\cite{wang2019learning} measures the similarity between region and expression embedding with an MLP, and the one-stage FAOA~\cite{yang2019fast} encodes the language vector to visual feature by direct concatenation.  These simple designs are efficient but lead to sub-optimal results, especially on long and complex language expressions. Following studies have proposed diverse architectures to improve the performance.
Among two-stage methods, modular attention network~\cite{yu2018mattnet}, various graphs~\cite{wang2019neighbourhood,yang2019dynamic,yang2020graph}, and multi-modal tree~\cite{liu2019learning} are designed to better model the multi-modal relationships.
The one-stage method~\cite{yang2020improving} has also explored better query modeling by proposing a multi-round fusion module.

Despite the effectiveness, these complicated fusion modules are built on certain pre-defined structures of language queries or image scenes, inspired by the human prior. Typically, the involvement of manually-designed mechanisms in fusion module makes the models overfit to specific scenarios, such as certain query lengths and query relationships, and \eat{thus potentially}limits the plenitudinous interaction between visual-linguistic contexts. \eat{and hurts the performance?} Moreover, even though the ultimate goal of visual grounding is to localize the referred object, most of the previous methods ground the queried object in an indirect fashion. 
They generally define surrogate problems of language-guided candidates prediction, selection, and refinement. Typically, the candidates are sparse region proposals~\cite{yu2016modeling,mao2016generation,wang2019learning} or dense anchors~\cite{yang2019fast}, from which the best region is selected and refined to get the final grounding box.
Since these methods' predictions are made out of candidates, the performance is easily influenced by the prior knowledge to generate proposals (or pre-defined anchors) and by the heuristics to assign targets to candidates. 

In this study, we explore an alternative approach to avoid the aforementioned problems. Formally, we introduce a neat and novel transformer-based framework, namely TransVG, to effectively address the task of visual grounding. We empirically show that the structurized fusion modules can be replaced by a simple stack of transformer encoder layers. Particularly, the core component of transformers (\ie, attention layer) is ready to establish intra-modality and inter-modality correspondence across visual and linguistic inputs, despite that we do not pre-define any specific fusion mechanism. Besides, we find that directly regressing the box coordinates works better than previous methods to ground the queried object indirectly. Our TransVG directly outputs 4-dim coordinates to ground the object instead of making predictions based on a set of candidate boxes.

The pipeline of our proposed TransVG is illustrated in Figure~\ref{fig:intro}(c). We first feed the RGB image and language expression into two sibling branches. The visual transformer and linguistic transformer are applied in these two branches to model the global cues in vision and language domains, respectively. Then, the abstracted visual tokens and linguistic tokens are fused, and the visual-linguistic transformer is exploited to perform cross-modal relation reasoning. Finally, the box coordinates of a referred object are directly regressed to make grounding. We benchmark our framework on five prevalent visual grounding datasets, including ReferItGame~\cite{kazemzadeh2014referitgame}, Flickr30K Entities~\cite{plummer2017flickr30k}, RefCOCO~\cite{yu2016modeling}, RefCOCO+~\cite{yu2016modeling}, RefCOCOg~\cite{mao2016generation}, and our method sets a series of state-of-the-art records. Remarkably, our proposed TransVG achieves
70.73\%, 79.10\% and 78.35\% on the test set of ReferItGame, Flickr30K and RefCOCO datasets, with 6.13\%, 5.80\%, 6.05\% absolute improvements over the strongest competitors.

In summary, we make three-fold contributions:
\begin{itemize}[noitemsep,nolistsep]
	\item We propose the first transformer-based framework for visual grounding, which holds neater architecture yet achieves better performance than the prevalent one-stage and two-stage frameworks.
	\item We present an elegant view of capturing intra- and inter-modality context homogeneously by transformers, and formulating visual grounding as a direct coordinates regression problem.
	\item We conduct extensive experiments to validate the merits of our method, and show significantly improved results on several prevalent benchmarks.
\end{itemize}

\section{Related Work}
\subsection{Visual Grounding}
Recent advances in visual grounding can be broadly categorized into two directions, \emph{i.e.}, two-stage methods~\cite{hong2019learning,hu2017modeling,liu2019learning,wang2019learning,wang2019neighbourhood,yang2019dynamic,yu2018mattnet,zhang2018grounding,zhuang2018parallel} and one-stage methods~\cite{chen2018real,liao2020real,sadhu2019zero,yang2020improving,yang2019fast}. We briefly review them in the following.

\textbf{Two-stage Methods.}
Two-stage approaches are characterized by generating region proposals in the first stage and then leveraging the language expression to select the best matching region in the second stage. 
Generally, the region proposals are generated using either unsupervised methods~\cite{plummerCITE2018,wang2019learning} or a pre-trained object detector~\cite{yu2018mattnet,zhang2018grounding}. The training loss of either binary classification~\cite{wang2019learning,zhang2017discriminative} or maximum-margin ranking~\cite{mao2016generation, nagaraja2016modeling, wang2016learning} is applied in the second stage to maximize the similarity between the positive object-query pair. Pioneer studies~\cite{mao2016generation,wang2016learning,yu2016modeling} obtain good results with the two-stage framework. The early work MattNet~\cite{yu2018mattnet} introduces the modular design and improves the grounding accuracy by better modeling the subject, location, and relation-related language description. Some recent studies further improve the two-stage methods by better modeling the object relationships~\cite{liu2019learning,wang2019neighbourhood,yang2019dynamic}, enforcing correspondence learning~\cite{liu2019improving}, or making use of phrase co-occurrences~\cite{bajaj2019g3raphground,chen2017query,dogan2019neural}.

\textbf{One-stage Methods.} One-stage approaches get rid of the computation-intensive object proposal generation and region feature extraction in the two-stage paradigm. Instead, the linguistic context is densely fused with the visual features, and the language-attended feature maps are further leveraged to perform bounding box prediction in a sliding-window manner. The pioneering work FAOA~\cite{yang2019fast} encodes the text expression into a language vector, and fuses the language vector into the YOLOv3 detector~\cite{redmon2018yolov3} to ground the referred instance. RCCF~\cite{liao2020real} formulates the visual grounding problem as a correlation filtering process~\cite{bolme2010visual,henriques2014high}, and picks the peak value of the correlation heatmap as the center of target objects. The recent work ReSC~\cite{yang2020improving}  devises a recursive sub-query construction module to address the limitations of FAOA~\cite{yang2019fast} on grounding complex queries.

\begin{figure}[t]
	\centering {\includegraphics[width=0.48\textwidth]{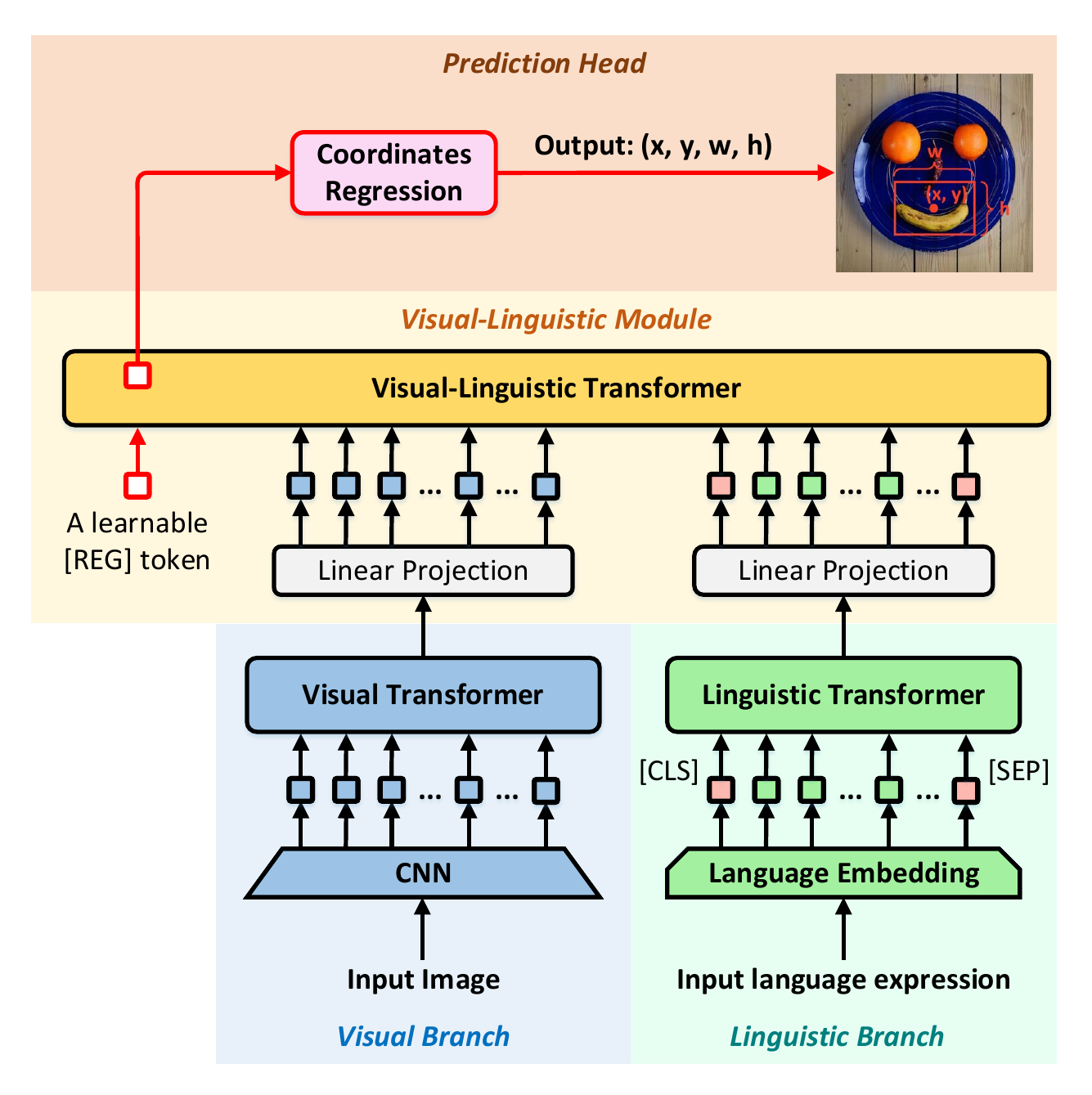}}
	
	\caption{An overview of our proposed TransVG framework. It consists of four main components: (1) a visual branch, (2) a linguistic branch, (3) a visual-linguistic fusion module, and (4) a prediction head to regress the box coordinates.} 
	\label{fig:framework}
\end{figure}

\subsection{Transformer}
Transformer is first proposed in~\cite{vaswani2017attention} to tackle the neural machine translation (NMT). The primary component of a transformer layer is the attention module, which scans through the input sequence in parallel and aggregates the information of the whole sequence with adaptive weights. Compared to the recurrent units in RNNs~\cite{hochreiter1997long,mikolov2010recurrent,tai2015improved}, the attention mechanism exhibits better performance in processing long sequences. This superiority has attracted a surge of research interest in applications of transformers in NLP tasks~\cite{dehghani2018universal,devlin2018bert,raffel2019exploring,zhu2020incorporating} and speech recognition~\cite{moritz2020streaming,wang2020transformer}.

\textbf{Transformer in Vision Tasks.} Inspired by the great success of transformers in neural machine translation, a series of  transformers~\cite{carion2020end,chen2020pre,chen2020generative,dosovitskiy2020image,huang2020hand,yang2020learning,zeng2020learning,zhu2020deformable} applied to vision tasks have been proposed. The infusive work DETR~\cite{carion2020end} formulates object detection as a set prediction problem. It introduces a small set of learnable object queries, reasons global context and object relations with attention mechanism, and outputs the final set of predictions in parallel. ViT~\cite{dosovitskiy2020image} shows that a pure transformer can achieve excellent performance on image classification tasks. More recently, a pre-trained image processing transformer (IPT) is introduced in~\cite{chen2020pre}  to address the low-level vision problems, \emph{e.g.}, denoising, super-resolution and deraining.

\textbf{Transformer in Vision-Language Tasks.} Motivated by the powerful pre-trained model of BERT~\cite{devlin2018bert}, some researchers start to investigate visual-linguistic pre-training (VLP) ~\cite{chen2020uniter,li2020oscar,lu2019vilbert,su2019vl,yang2020tap} to jointly represent images and texts. In general, these models take the object proposals and text as inputs, and devise several transformer encoder layers for joint representation learning. Plenty of pre-training tasks are introduced, including image-text matching (ITM), word-region alignment (WRA), masked language modeling (MLM), masked region modeling (MRM), \emph{etc.}

Although with similar base units (\ie transformer encoder layers), the goal of VLP is to learn a generalizable vision-language representation with large-scale data to facilitate down-stream tasks. In contrast, we focus on developing a novel transformer-based visual grounding framework, and learning to perform homogeneous multi-modal  reasoning with a small amount of visual grounding data.

\section{Transformers for Visual Grounding}
In this work, we present Transformers for Visual Grounding (TransVG), a novel framework for the visual grounding task based on a stack of transformer encoders with direct box coordinates prediction. As shown in Figure~\ref{fig:framework}, given an image and a language expression as inputs, we first separate them into two sibling branches, \emph{i.e.}, a visual branch and a linguistic branch, to generate visual and linguistic feature embedding. Then, we put the multi-modal feature embedding together and append a learnable token (named [REG] token) to construct the inputs of visual-linguistic fusion modules. The visual-linguistic transformer homogeneously embeds the input tokens from different modalities into a common semantic space by modeling intra-modality and inter-modality context with the self-attention mechanism. Finally, the output state of the [REG] token is leveraged to directly predict the 4-dim coordinates of a referred object in the prediction head. 

In the following subsections, we first review the preliminary for transformer and then elaborate our designs of transformers for visual grounding.

\subsection{Preliminary}
Before detailing the architecture of TransVG, we briefly review the conventional transformer proposed in~\cite{vaswani2017attention} for machine translation. The core component in a transformer is the attention mechanism. Given the query embedding $\bm{f}^q$, key embedding $\bm{f}^k$ and value embedding $\bm{f}^v$, the output of a single-head attention layer is computed as:
\begin{equation}
	\label{func:attn}
	\text{Attn}(\bm{f}^q,\bm{f}^k,\bm{f}^v) = \text{softmax}(\frac{\bm{f}^q\bm{f}^k}{\sqrt{{d}^k}})\cdot\bm{f}^v,
\end{equation}
where $d^k$ is the channel dimension of $\bm{f}^k$. Similar to classic neural sequence transduction models, the conventional transformer has an encoder-decoder structure. However, in our approach, we only use transformer encoder layers.

Concretely, each transformer encoder layer has two sub-layers, \emph{i.e.}, a multi-head self-attention layer and a simple feed forward network (FFN). The multi-head attention is a variant of single-head attention (as in Function~\ref{func:attn}), and self-attention indicates the query, key and value are from the same embedding set. FFN is an MLP composed of fully connected layers and ReLU activation layers. 

In the transformer encoder layer, each sub-layer is put into a residual structure, where layer normalization~\cite{ba2016layer} is performed after the residual connection. 
Let us denote the input as $\bm{x}_n$, the procedure in a transformer encoder layer is:
\begin{align}
	\bm{x}'_n &= \text{LN}(\bm{x}_n + \mathcal{F}_\text{MSA}(\bm{x}_n)), \label{eq:msa} \\
	\bm{x}_{n+1} &= \text{LN}(\bm{x}'_n + \mathcal{F}_\text{FFN}(\bm{x}'_n)), \label{eq:FFN}
\end{align}
where $\text{LN}(\cdot)$ indicates layer normalization, $\mathcal{F}_\text{MSA}(\cdot)$ is the multi-head self-attention layer, and $\mathcal{F}_\text{FFN}(\cdot)$ represents the feed forward network.

\subsection{TransVG Architecture}
As depicted in Figure~\ref{fig:framework}, there are four main components in TransVG: (1) a visual branch, (2) a linguistic branch, (3) a visual-linguistic fusion module, and (4) a prediction head. 

\noindent{\textbf{Visual Branch.}} 
The visual branch starts with a convolutional backbone network, followed by the visual transformer. We exploit the commonly used ResNet~\cite{he2016deep} as the backbone network. The visual transformer is composed of a stack of 6 transformer encoder layers. Each transformer encoder layer includes a multi-head self-attention layer and an FFN. There are 8 heads in the multi-head attention layer, and 2 FC layers followed by ReLU activation layers in the FFN. The output channel dimensions of these 2 FC layers are 2048 and 256, respectively.

Given an image $\bm{z}_\text{0}\in\mathbb{R}^{3\times H_0 \times W_0}$ as the input of this branch, we exploit the backbone network to generate a 2D feature map $\bm{z} \in\mathbb{R}^{C\times H \times W}$. Typically, the channel dimension $C$ is $2048$, and the width and height of the 2D feature map are $\frac{1}{32}$ of the original image size ($H=\frac{H_0}{32}$, $W=\frac{W_0}{32}$). Then, we leverage a $1\times1$ convolutional layer to reduce the channel dimension of $\bm{z}$ to $C_v=256 $ and obtain $\bm{z}' \in \mathbb{R}^{C_v \times H \times W}$. Since the input of a transformer encoder layer is expected to be a sequence of 1D vectors, we further flatten $\bm{z}'$ into $\bm{z}_v \in \mathbb{R}^{C_v\times N_v}$, where $N_v=H\times W$ is the number of input tokens. To make the visual transformer sensitive to the original 2D positions of input tokens, we follow~\cite{carion2020end,parmar2018image} to utilize sine spatial position encodings as the supplementary of visual feature. Concretely, the position encodings are added with the query and key embedding at each transformer encoder layer. The visual transformer conducts global vision context reasoning in parallel, and outputs the advanced visual embedding $\bm{f}_v$, which shares the same shape as $\bm{z}_v$.

\noindent{\textbf{Linguistic Branch.}} 
The linguistic branch is a sibling to the visual branch. Our linguistic branch includes a token embedding layer and a linguistic transformer. To make the best of the pre-trained BERT model~\cite{devlin2018bert}, the architecture of this branch follows the design of the basic model of BERT series. Typically, there are 12 transformer encoder layers in the linguistic transformer. The output channel dimension of the linguistic transformer is $C_l=768$.

Given a language expression as the input of this branch, We first convert each word ID into a one-hot vector. Then, in the token embedding layer, we tokenize each one-hot vector into a language token by looking up the token table. We follow the common practice in machine translation~\cite{dehghani2018universal,devlin2018bert,raffel2019exploring,vaswani2017attention} to append a [CLS] token and a [SEP] token at the beginning and end positions of the tokenized language expression. After that, we take the language tokens as inputs of the linguistic transformer, and generate the advanced language embedding $\bm{f}_l \in \mathbb{R}^{C_l \times N_l}$, where $N_l$ is the number of language tokens. 

\noindent{\textbf{Visual-linguistic Fusion Module.}}
As the core component in our model to fuse the multi-modal context, the architecture of the visual-linguistic fusion module (abbreviated as V-L module) is simple and elegant. Specifically, the V-L module includes two linear projection layers (one for each modality) and a visual-linguistic transformer (with a stack of 6 transformer encoder layers).

Given advanced visual tokens $\bm{f}_v\in\mathbb{R}^{256\times N_v}$ out of the visual branch and advanced linguistic tokens $\bm{f}_l\in\mathbb{R}^{768\times N_l}$ out of the linguistic branch, we apply a linear projection layer to project them into embedding with same channel dimension. We denote the projected visual embedding and linguistic embedding as $\bm{p}_v\in{\mathbb{R}^{C_p\times N_v}}$ and $\bm{p}_l\in{\mathbb{R}^{C_p\times N_l}}$, where $C_p=256$. Then, we pre-append a learnable embedding (namely a [REG] token) to $\bm{p}_v$ and $\bm{p}_l$, and formulate the joint input tokens of the visual-linguistic transformer as:
\begin{equation}
	\label{eq:token}
	\bm{x}_0 = [\ \underbrace{p_v^1,\ p_v^2,\ \cdots,\ p_v^{N_v}}_{\text{visual tokens}~\bm{p}_v},\ \overbrace{p_l^1,\ p_l^2,\ \cdots,\ p_l^{N_l}}^{\text{linguistic tokens}~\bm{p}_l},\ p_r\ ],
\end{equation}
where $p_r\in\mathbb{R}^{C_p \times1}$ represents the [REG] token. The [REG] token is randomly initialized at the beginning of the training stage and optimized with the whole model.

After obtaining the input $\bm{x}_0\in\mathbb{R}^{C_p\times(N_v+N_l+1)}$ in the joint embedding space as described above, we apply the visual-linguistic transformer to embed $\bm{x}_0$ into a common semantic space by performing intra- and inter-modality relation reasoning in a homogeneous way. To retain the positional and modal information, we add learnable position encodings to the input of each transformer encoder layer.

Thanks to the attention mechanism, the correspondence can be freely established between each pair of tokens from the joint entities, regardless of their modality. For example, a visual token can attend to a visual token, and it can also freely attend to a linguistic token. Typically, the output state of the [REG] token develops a consolidated representation enriched by both visual and linguistic context, and is further leveraged for box coordinates prediction.

\noindent{\textbf{Prediction Head.}}
We leverage the output state of [REG] token from the V-L module as the input of our prediction head. To perform box coordinates prediction, we attach a regression block to the [REG] token. The regression block is implemented by an MLP with two ReLU activated 256-dim hidden layers and a linear output layer. The output of the prediction head is the 4-dim box coordinates.

\subsection{Training Objective}
Unlike many previous methods that ground referred objects based on a set of candidates (\emph{i.e.}, region proposals in two-stage methods and anchor boxes in one-stage methods), TransVG directly infers a 4-dim vector as the coordinates of the box to be grounded. This simplifies the process of target assignment and positive/negative examples mining at the training stage, but it also involves the scale problem. Specifically, the widely used smooth L1 loss tends to be a large number when we try to predict a large box, while tends to be small when we try to predict a small one, even if their predictions have similar relative errors. 

To address this problem, we normalize the coordinates of the ground-truth box by the scale of the image, and involve the generalized IoU loss~\cite{rezatofighi2019generalized} (GIoU loss), which is not affected by the scales.

Let us denote the prediction as $\bm{b}=(x, y, w, h)$, and the normalized ground-truth box as $\hat{\bm{b}}=(\hat{x},\hat{y},\hat{w},\hat{h})$. The training objective of our TransVG is:
\begin{equation}
	\label{func:loss}
	\mathcal{L} = \mathcal{L}_{\text{smooth-l1}}(\bm{b},\hat{\bm{b}}) + \lambda\cdot\mathcal{L}_{\text{giou}}(\bm{b},\hat{\bm{b}}),
\end{equation}
where $\mathcal{L}_{\text{smooth-l1}}(\cdot)$ and $\mathcal{L}_{\text{giou}}(\cdot)$ are the smooth L1 loss and GIoU loss, respectively. $\lambda$ is the weight coefficient of GIoU loss to balance these two losses.

\section{Experiments}

\subsection{Datasets}
\noindent{\bf ReferItGame.} ReferItGame~\cite{kazemzadeh2014referitgame}  includes 20,000 images collected from the SAIAPR-12 dataset~\cite{escalante2010segmented}, and each image has one or a few regions with  corresponding referring expressions. We follow the common practice to divide this dataset into three subsets, \emph{i.e.}, a train set with 54,127 referring expressions, a validation set with 5,842 referring expressions and a test set with 60,103 referring expressions. We use the validation set to conduct experimental analysis and compare our method with others on the test set.

\noindent{\bf Flickr30K Entities.} Flickr30K Entities~\cite{plummer2017flickr30k} augments the original Flickr30K \cite{young2014image} with short region phrase correspondence annotations. It contains 31,783 images with 427K referred entities. We follow the previous works~\cite{plummer2017flickr30k,plummerCITE2018,wang2019learning,yang2020improving} to separate the these images into 29,783 for training, 1000 for validation, and 1000 for testing.


\noindent{\bf RefCOCO/ RefCOCO+/ RefCOCOg.} 
RefCOCO~\cite{yu2016modeling} includes 19,994 images with 50,000 referred objects. Each object has more than one referring expression, and there are 142,210 referring expressions in this dataset. The samples in RefCOCO are officially split into a train set with 120,624 expressions, a validation set with 10,834 expressions, a testA set with 5,657 expressions and a testB set with 5,095 expressions. Similarly, RefCOCO+~\cite{yu2016modeling} contains 19,992 images with 49,856 referred objects and 141,564 referring expressions. It is also officially split into a train set with 120,191 expressions, a validation set with 10,758 expressions, a testA set with 5,726 expressions and a testB set with 4,889 expressions. RefCOCOg~\cite{mao2016generation} has 25,799 images with 49,856 referred objects and expressions. There are two commonly used split protocols for this dataset. One is RefCOCOg-google~\cite{mao2016generation}, and the other is RefCOCOg-umd~\cite{nagaraja2016modeling}. We report our performance on both RefCOCOg-google (val-g) and RefCOCOg-umd (val-u and test-u) to make comprehensive comparisons.

\subsection{Implementation Details}
\noindent{\textbf{Inputs.}} We set the input image size as $640\times640$ and the max expression length as 40. When performing image resizing, we keep the original aspect ratio of each image. The longer edge of an image is resized to 640, while the shorter one is padded to 640 with the mean value of RGB channels. Meanwhile, We cut off the language query if its length is longer than 38 (leaving one position for the [CLS] token and one position for the [SEP] token). Otherwise, we pad empty tokens after [SEP] token to make the input length equal to 40. For both the input image and language expression, the padded pixel/word is recorded with a mask and will not be involved in the computation of transformers. 

\noindent{\textbf{Training Details.}} The whole architecture of our TransVG is end-to-end optimized with AdamW optimizer. We set the initial learning rate of the V-L module and prediction head to $10^{-4}$, the visual branch and linguistic branch to $10^{-5}$, and set weight decay to $10^{-4}$. Our visual branch is initialized with the backbone and encoder of DETR model~\cite{carion2020end}, and our linguistic branch is initialized with the basic BERT model~\cite{devlin2018bert}. For the other components, the parameters are randomly initialized with Xavier init. On all the datasets except Flickr30K Entities, our model is trained for 90 epochs with a learning rate dropped by a factor of 10 after 60 epochs. As for the Flickr30K Entities, our model is trained for 60 epochs, with a learning rate drops after 40 epochs. We set the batch size to 64. The weight coefficient $\lambda$ is set to 1. To avoid overfitting, we exploit dropout operation after the multi-head self-attention layer and the FFN of each transformer encoder layer. The dropout ratio is set to 0.1 by default. We follow the common practice in ~\cite{liao2020real,yang2020improving,yang2019fast} to perform data augmentation at the training stage.

\noindent{\textbf{Inference.}} Since our TransVG directly outputs the box coordinates, there is no extra operation at the inference stage.

\subsection{Comparisons with State-of-the-art Methods}
To validate the merits of our proposed TransVG, we report our performance and compare it with other state-of-the-art methods on five visual grounding benchmarks, including ReferItGame~\cite{kazemzadeh2014referitgame}, Flickr30K Entities~\cite{plummer2017flickr30k}, RefCOCO~\cite{yu2016modeling}, RefCOCO+~\cite{yu2016modeling}, and RefCOCOg~\cite{mao2016generation}. We follow the standard protocol to report the performance in terms of top-1 accuracy (\%). Specifically, once the Jaccard overlap between the predicted region and the ground-truth box is above 0.5, the prediction is regarded as a correct one.

\noindent{\textbf{ReferItGame.}} 
Table~\ref{tab:referit_results} shows the result comparison between state-of-the-art methods on the ReferItGame test set. We group the methods into two-stage methods, one-stage methods, and transformer-based methods. Among all the methods, TransVG achieves the best performance as the first transformer-based approach. With ResNet-50 backbone, TransVG achieves 69.76\% top-1 accuracy and outperforms ZSGNet~\cite{sadhu2019zero} with the same backbone network by 11.13\%. By replacing ResNet-50 with a stronger ResNet-101, the performance boosts to 70.73\%, which is 6.13\% higher than the strongest competitor ReSC-Large for one-stage methods and 7.73\% higher than the strongest competitor DDPN for two-stage methods, respectively. 

In particular, we find the recurrent architecture in ReSC shares the same spirit with our stacking architecture in the visual-linguistic transformer that fuses the multi-modal context in multiple rounds. However, in ReSC, recurrent learning is only performed to construct the language sub-query, and this procedure is isolated from the sub-query attended visual feature modulation. In contrast, our TransVG embeds the visual and linguistic embedding into a common semantic space by homogeneously performing intra- and inter-modality context reasoning. The superiority of our performance empirically demonstrates the effectiveness of our unified visual-linguistic encoder and fusion module designs. It also validates that the complicated multi-modality fusion module can be replaced by a simple stack of transformer encoder layers.

\begin{table}[t]
	\caption{Comparisons with state-of-the-art methods on the test set of  ReferItGame~\cite{kazemzadeh2014referitgame} and Flickr30K Entities~\cite{plummer2017flickr30k} in terms of top-1 accuracy (\%). The previous methods follow the two-stage or one-stage directions, while ours is transformer-based. We highlight the best and second best performance in the \textcolor{red}{red} and \textcolor{blue}{blue} colors.}
	\small
	\begin{center}
		\scalebox{0.9}{
			\begin{tabular}{c | c | c | c  }
				\hline
				\multirow{2}{*}{Models} & \multirow{2}{*}{Backbone} & \multicolumn{1}{c|}{ReferItGame} & \multicolumn{1}{c}{Flickr30K}\\
				& & test & test\\
				\hline
				\hline
				\textbf{\textit{Two-stage:}}  &  &  &\\
				CMN~\cite{hu2017modeling} & VGG16 & 28.33 & -\\
				VC~\cite{zhang2018grounding} & VGG16 & 31.13 & - \\
				MAttNet~\cite{yu2018mattnet} &	ResNet-101  & 29.04 & - \\
				Similarity Net~\cite{wang2019learning} & ResNet-101 & 34.54 & 60.89\\
				CITE~\cite{plummerCITE2018}  & ResNet-101  & 35.07 & 61.33\\
				PIRC~\cite{kovvuri2018pirc} & ResNet-101 & 59.13 & 72.83 \\
				DDPN~\cite{yu2018rethinking} & ResNet-101 &  63.00 & 73.30\\
				\hline
				\textbf{\textit{One-stage:}}  &  & &\\
				SSG~\cite{chen2018real} & DarkNet-53 & 54.24 & - \\
				ZSGNet~\cite{sadhu2019zero} & ResNet-50 & 58.63 & 63.39\\
				FAOA~\cite{yang2019fast} & DarkNet-53 & 60.67 & 68.71\\
				RCCF~\cite{liao2020real} & DLA-34 & 63.79 & -\\
				ReSC-Large~\cite{yang2020improving} & DarkNet-53 & 64.60 & 69.28\\
				
				\hline
				\textbf{\textit{Transformer-based:}}  &  &   &\\
				TransVG (ours) & ResNet-50 &  \bm{\textcolor{blue}{69.76}} & \bm{\textcolor{blue}{78.47}} \\
				TransVG (ours) & ResNet-101 &  \bm{\textcolor{red}{70.73}} & \bm{\textcolor{red}{79.10}} \\
				
				\hline
			\end{tabular}
		} 
	\end{center}
	\label{tab:referit_results}
	\vspace{-0.10cm}
\end{table}

\noindent{\textbf{Flickr30K Entities.}} 
Table~\ref{tab:referit_results} also reports the performance of our TransVG on the Flickr30K Entities test set. On this dataset, our TransVG achieves 79.10\% top-1 accuracy with a ResNet-101 backbone network, surpassing the recently proposed Similarity Net~\cite{wang2019learning}, CITE~\cite{plummerCITE2018}, DDPN~\cite{yu2018rethinking}, ZSGNet~\cite{sadhu2019zero}, FAOA~\cite{yang2019fast}, and ReSC-Large~\cite{yang2020improving} by a remarkable margin (\emph{i.e.}, 5.80\% absolute improvement over the previous state-of-the-art record).

\begin{table*}[t]
	\caption{Comparisons with state-of-the-art methods on RefCOCO~\cite{yu2016modeling}, RefCOCO+~\cite{yu2016modeling} and RefCOCOg~\cite{mao2016generation} in terms of top-1 accuracy (\%). We highlight the best and second best performance in the \textcolor{red}{red} and \textcolor{blue}{blue} colors. We have added the reproduced performance with our released code, denoted as TransVG (released) in the table. The following works are encouraged to compare with the reproduced performance instead of the original one.}
	
	\vspace{-0.2cm}
	
	\small
	\begin{center}
		\scalebox{0.8}[0.8]{
			\setlength
			\tabcolsep{9.4pt}
			\begin{tabular}{c | c | c | c c c | c c c | c c c }
				\hline
				\multirow{2}{*}{Models} & \multirow{2}{*}{Venue} & \multirow{2}{*}{Backbone} & \multicolumn{3}{c|}{RefCOCO} & \multicolumn{3}{c|}{RefCOCO+} & \multicolumn{3}{c}{RefCOCOg} \\ 
				
				&  &  & val & testA & testB & val & testA & testB & val-g & val-u & test-u \\
				\hline \hline
				\textbf{\textit{Two-stage:}} & &  & & & & & & & & &\\
				CMN~\cite{hu2017modeling} & \textit{CVPR'17} & VGG16 & - & 71.03 & 65.77 & - & 54.32 & 47.76 & 57.47 & - & - \\
				VC~\cite{zhang2018grounding} & \textit{CVPR'18} & VGG16 & - & 73.33 & 67.44 & - & 58.40 & 53.18 &62.30 & - & - \\
				ParalAttn~\cite{zhuang2018parallel} & \textit{CVPR'18} & VGG16 & - & 75.31 & 65.52 & - & 61.34 & 50.86 & 58.03 & - & - \\
				MAttNet~\cite{yu2018mattnet} & \textit{CVPR'18} &	ResNet-101 & 76.65 & 81.14 & 69.99 & \bm{\textcolor{blue}{65.33}} & \bm{\textcolor{blue}{71.62}} & 56.02 & -  & 66.58 & 67.27 \\
				LGRANs~\cite{wang2019neighbourhood} & \textit{CVPR'19} & VGG16 & - & 76.60 & 66.40 & - & 64.00 & 53.40 & 61.78 & - & - \\
				DGA~\cite{yang2019dynamic} & \textit{ICCV'19} & VGG16 & - & 78.42 & 65.53 & - & 69.07 & 51.99 & - & - & 63.28 \\ 
				RvG-Tree~\cite{hong2019learning} & \textit{TPAMI'19} & ResNet-101 & 75.06 & 78.61 & 69.85 & 63.51 & 67.45 & 56.66 & - & 66.95 & 66.51 \\
				NMTree~\cite{liu2019learning} & \textit{ICCV'19} & ResNet-101 & 76.41 & 81.21 & 70.09 & \bm{\textcolor{red}{66.46}} & \bm{\textcolor{red}{72.02}} & \bm{\textcolor{red}{57.52}} & 64.62 & 65.87 & 66.44 \\
				\hline
				\textbf{\textit{One-stage:}} & &  & & & & & & & & &\\
				SSG~\cite{chen2018real} & \textit{arXiv'18} &  DarkNet-53 & - & 76.51 & 67.50 & - & 62.14 & 49.27 & 47.47 & 58.80 & - \\  
				FAOA~\cite{yang2019fast} & \textit{ICCV'19} & DarkNet-53 & 72.54 & 74.35 & 68.50 & 56.81 & 60.23 & 49.60 & 56.12 & 61.33 & 60.36 \\
				RCCF~\cite{liao2020real} & \textit{CVPR'20} & DLA-34 & - & 81.06 & 71.85 & - & 70.35 & 56.32 & -  & - & 65.73 \\
				ReSC-Large~\cite{yang2020improving} & \textit{ECCV'20} & DarkNet-53 & 77.63 & 80.45 & 72.30 & 63.59 & 68.36 & 56.81 & 63.12 & 67.30 & 67.20 \\
				\hline
				\textbf{\textit{Transformer-based:}} & &  & & & & & & & & &\\
				TransVG (ours) & - & ResNet-50 & \bm{\textcolor{blue}{80.32}} & \bm{\textcolor{blue}{82.67}} & \bm{\textcolor{blue}{78.12}} & 63.50 & 68.15 & 55.63 & \bm{\textcolor{blue}{66.56}} & \bm{\textcolor{blue}{67.66}} & \bm{\textcolor{blue}{67.44}} \\
				TransVG (ours) & - & ResNet-101 & \bm{\textcolor{red}{81.02}} & \bm{\textcolor{red}{82.72}} & \bm{\textcolor{red}{78.35}} & 64.82 & 70.70 & \bm{\textcolor{blue}{56.94}} & \bm{\textcolor{red}{67.02}} & \bm{\textcolor{red}{68.67}} & \bm{\textcolor{red}{67.73}} \\
				\hline
				TransVG (released)& - & ResNet-50 & 80.49 & 83.28 & 75.24 & 66.39 & 70.55 & 57.66 & 66.35 & 67.93 & 67.44 \\
				TransVG (released)& - & ResNet-101 & 80.83 & 83.38 & 76.94 & 68.00 & 72.46 & 59.24 & 68.03 & 68.71 & 67.98 \\
				\hline
			\end{tabular}
		} 
	\end{center}
	\label{tab:refcoco_results}
	\vspace{-0.5cm}
\end{table*}

\noindent{\textbf{RefCOCO/RefCOCO+/RefCOCOg.}}
To further validate the effectiveness of our proposed TransVG, we also conduct experiments to report our performance on the RefCOCO, RefCOCO+ and RefCOCOg datasets. The top-1 accuracy (\%) of our method, together with other state-of-the-art methods, is reported in Table~\ref{tab:refcoco_results}. Our TransVG consistently achieves the best performance on the RefCOCO and RefCOCOg for all the subsets and splits. Remarkably, we achieve 78.35\% on the RefCOCO testB set, 6.05\% absolute improvement over the previous state-of-the-art result. When performing grounding on longer expressions (on the RefCOCOg dataset), our method also works well, which further validates our neat architecture's effectiveness to process complicated queries. On RefCOCO+, TransVG also achieves comparable performance to that with the best records. We study the failure cases and find some extreme examples whose expressions are not suitable for generating embedding with transformers. For example, a query that just tells a number ``32'' in the annotation degenerates our linguistic transformer to an MLP in this situation. 

Among the competitors, MAttNet~\cite{yu2018mattnet} is the most representative method that devises multi-modal fusion modules with re-defined structures (\emph{i.e}, modular attention networks to separately model subject, location and relationship). When we compare our model with MAttNet in Table~\ref{tab:referit_results} and Table~\ref{tab:refcoco_results}, we can find that MAttNet shows comparable results to our TransVG on RefCOCO/RefCOCO+/RefCOCOg, but lags behind our TransVG on RefeItGame. The reason is that the pre-defined relationship in multi-modal fusion modules makes it easy to overfit to specific datasets (\emph{e.g.}, with  specific  scenarios,  query  lengths,  and  relationships). Our TransVG theoretically avoids this problem by establishing intra-modality and inter-modality correspondence with the flexible and adaptive attention mechanism.

\subsection{Ablation Study}
In this section, we conduct ablative experiments to verify the effectiveness of each component in our proposed framework. We exploit ResNet-50 as the backbone network of the visual branch, and all of the compared models are trained for 90 epochs as described in the implementation details.

\noindent{\textbf{Design of the [REG] Token.}}
We study the design of the [REG] token on RefCOCO dataset, and report the results in Table~\ref{tab:ablative_token}. There are several choices to construct the initial state of the [REG] token. We detail these designs and analysis them as follows:
\begin{itemize}[nolistsep]
	\item[---] \textit{Average pooled visual tokens.} We perform average pooling over the visual tokens and exploit the average-pooled embedding as the initial state of [REG] token.
	\item[---] \textit{Max pooled visual tokens.} We take the max-pooled visual token embedding as the initial [REG] token.
	\item[---] \textit{Average pooled linguistic tokens.} Similar to the first choice, but using the linguistic tokens.
	\item[---] \textit{Average pooled linguistic tokens.} Similar to the second choice, but using the linguistic tokens.
	\item[---] \textit{Sharing with [CLS] token.} We use the [CLS] token of linguistic embedding to pl the [REG] token. Concretely, the [CLS] token out of the V-L module is fed into the prediction head.
	\item[---] \textit{Learnable embedding*.} This is our default setting by randomly initializing the [REG] token embedding at the beginning of the training stage. And the parameters of this embedding are optimized with the whole model.
\end{itemize}

Our proposed design to exploit a learnable embedding achieves 80.32\% top-1 accuracy on the validation set of RefCOCO, which is the best performance among all the designs. Typically, the initial [REG] token of other designs is either generated from visual or linguistic tokens, which involves biases to the specific prior context of the corresponding modality. In contrast, the learnable embedding tends to be more equitable and flexible when performing relation reasoning in the visual-linguistic transformer.

\begin{table}
	\centering
	\footnotesize
	\caption{Ablative experiments on RefCOCO to study the [REG] token design in our framework. The initial state of the [REG] token is either obtained from visual/linguistic tokens out of the corresponding branch or by exploiting a learnable embedding.}
	\renewcommand\arraystretch{1}
	\begin{center}
		\setlength\tabcolsep{13pt}
		\begin{tabular}{lc}
			\hline
			Initial State of [REG] Token & RefCOCO@val\\
			\hline
			Average pooled visual tokens & 79.12 \\
			Max pooled visual tokens & 78.37\\
			Average pooled linguistic tokens & 78.51 \\
			Max pooled linguistic tokens & 78.74 \\
			Sharing with [CLS] token & 77.90 \\
			
			$\text{Learnable embedding}^*$ & $\bm{80.32}$\\
			\hline
		\end{tabular}
		
	\end{center}
	\label{tab:ablative_token}
	\vspace{-0.55cm}
\end{table}

\noindent{\textbf{Transformers in Visual and Linguistic Branches.}}
We study the role of the transformers in the visual branch and the linguistic branch (\emph{i.e.}, visual transformer and linguistic transformer). Table~\ref{tab:ablative_tr} summarizes the results of several models with or without the visual transformer and the linguistic transformer. The baseline model without both visual transformer and linguistic transformer reports an accuracy of 64.24\%. When we only attach either the visual transformer or the linguistic transformer, an improvement of 68.48\% and 66.78\% are achieved, respectively. With the complete architecture, the performance is further boosted to 69.76\% on the ReferIt test set. This result demonstrates the essential of transformers in the visual branch and linguistic branch to capture intra-modality global context before performing multi-modal fusion.

\begin{figure}[t]
	\centering {\includegraphics[width=0.47\textwidth]{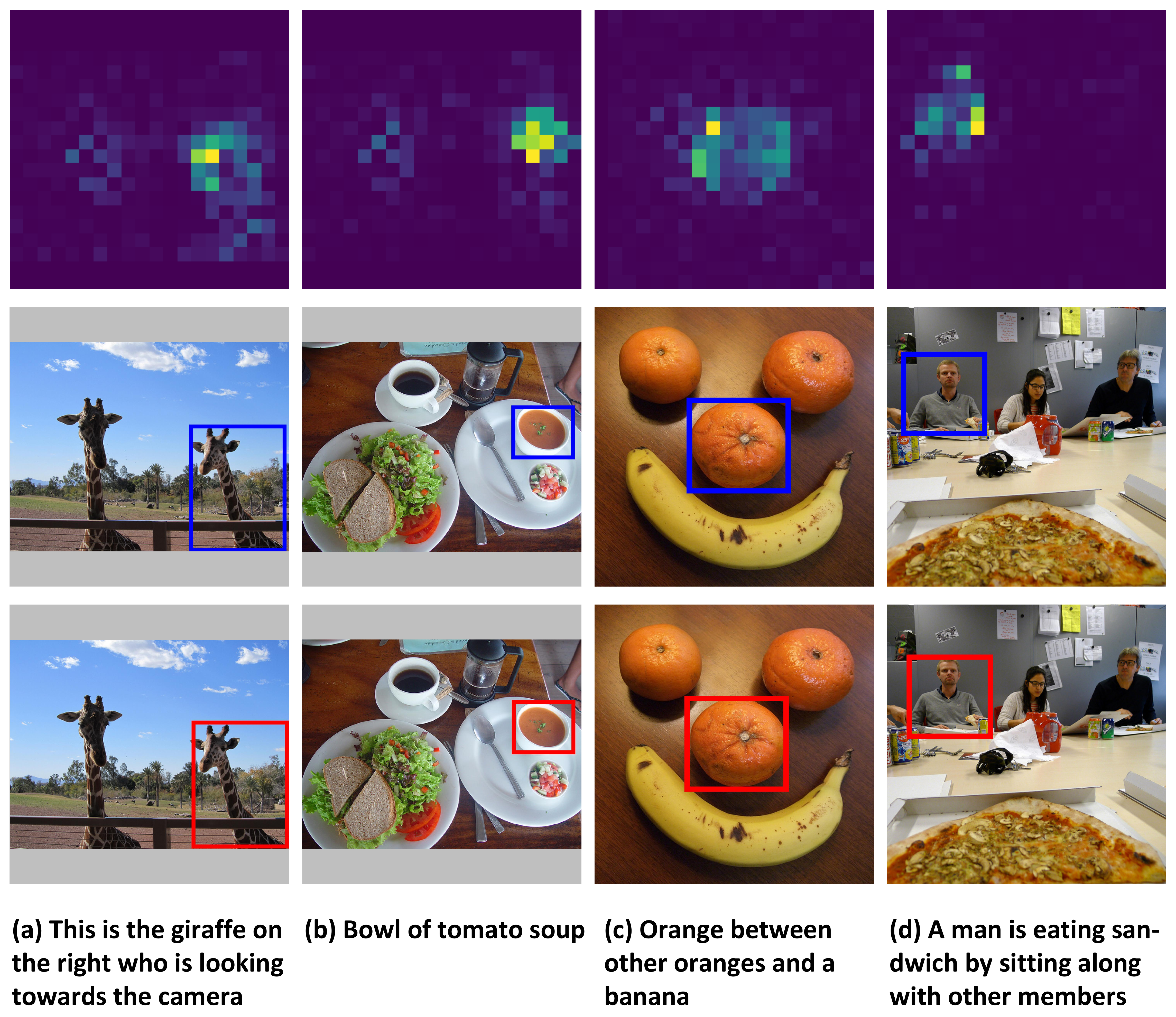}}
	\caption{Qualitative results of our TransVG on the RefCOCOg test set (better viewed in color). We show the [REG] token's attention to visual tokens in the top row. \textcolor{blue}{Blue} and \textcolor{red}{red} boxes are the predicted regions and the ground truths, respectively.}
	\label{fig:example}
	\vspace{-0.1in}
\end{figure}

\subsection{Qualitative Results}

We showcase the qualitative results of four examples from the RefCOCOg~\cite{mao2016generation} test set in Figure~\ref{fig:example}. We observe that our approach can successfully model queries with complicated relationships, \eg, ``orange between other oranges and a banana'' in Figure~\ref{fig:example}~(c). The first row of Figure~\ref{fig:example} visualizes the [REG] token's attention to the visual tokens in the visual-linguistic transformer. TransVG generates interpretable attentions on the referred object that corresponds to the overall object shape and location.

\begin{figure}[t]
	\centering {\includegraphics[width=0.47\textwidth]{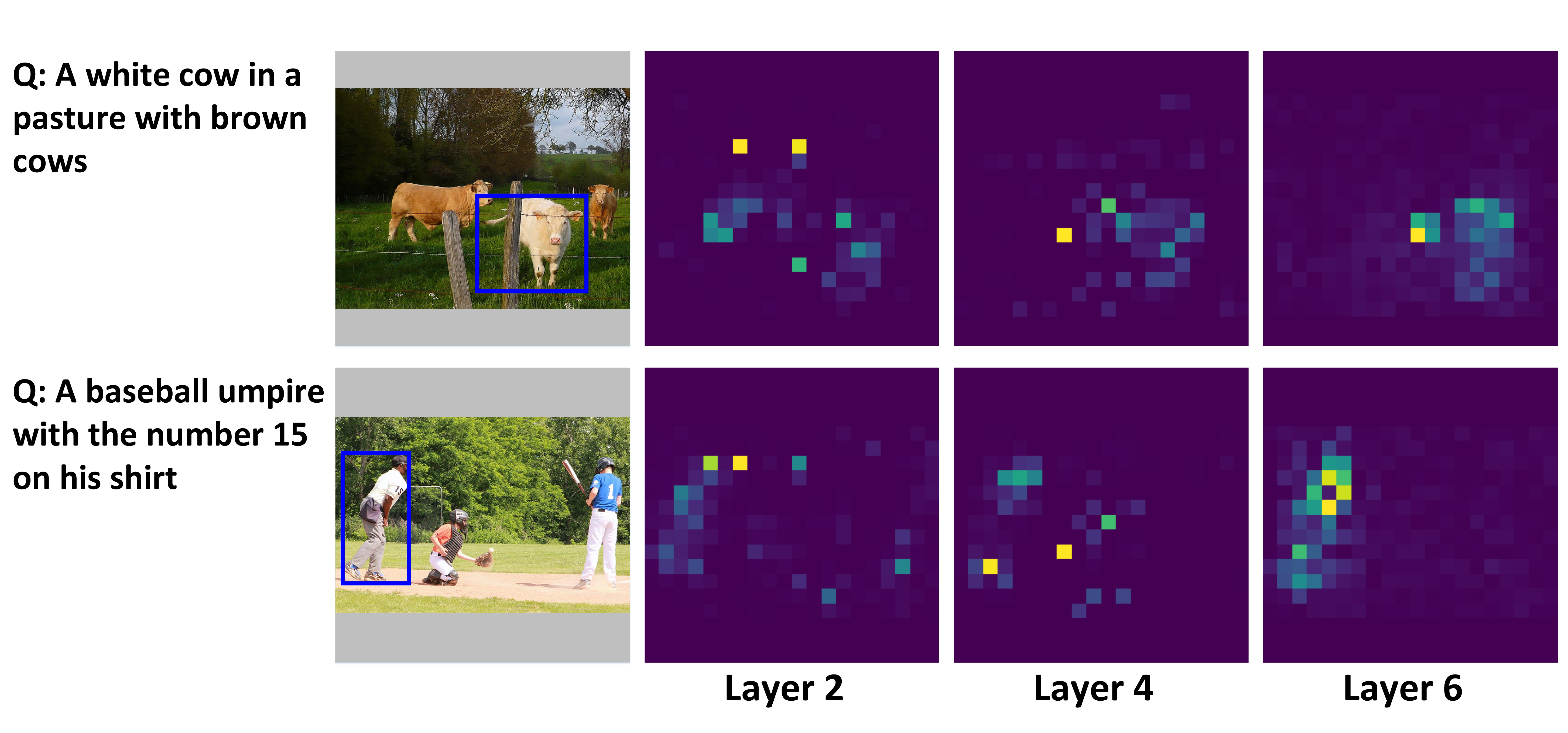}}
	\caption{Visualization of the [REG] token's attention score on visual tokens from the second (layer $2$), forth (layer $4$) and sixth (layer $6$) encoder layer of the visual-linguistic transformer.}
	\label{fig:attn}
	\vspace{-0.1in}
\end{figure}

Motivated by the correspondence between visual attention and predicted regions, we visualize the [REG] token's attention score on the visual tokens in the visual-linguistic transformer's intermediate layers to better understand TransVG. Figure~\ref{fig:attn} shows the [REG] token's attention score on the visual tokens from the second, forth and sixth transformer encoder layers. In the early layer (layer $2$), we observe that the [REG] token captures the global context by attending to multiple regions in the whole image. In the middle layer (layer $4$), the [REG] token tends to attend the discriminative regions which are closely related to the referred object (\emph{e.g.}, the bus behind the man in the first example, which indicates the scene is on the road). In the final layer (layer $6$), TransVG attends to the referred object and generates a more accurate attention prediction for the object's shape, which enables the model to regress the target's coordinates correctly.

\begin{table}
	\centering
	\footnotesize
	\caption{Ablative experiments of the visual transformer and linguistic transformer in our framework. The performance is evaluated on the test set of ReferItGame~\cite{kazemzadeh2014referitgame} in terms of top-1 accuracy (\%). ``Tr.'' represents transformer.}
	\renewcommand\arraystretch{1}
	\begin{center}
		\scalebox{0.95}[0.95]{
			\setlength
			\tabcolsep{7.2pt}
			\begin{tabular}{cc|cc|cc}
				\hline
				\multicolumn{2}{c|}{Visual Branch} & \multicolumn{2}{c|}{Linguistic Branch} & Accuracy & Runtime \\
				w/o Tr.  & w/ Tr. & w/o Tr. & w/ Tr. & (\%) & (ms) \\
				\hline
				\checkmark & & \checkmark &  & 64.24& 33.67\\
				\checkmark & &  &\checkmark  & $66.78_{\uparrow 3.54}$ & 47.57 \\
				& \checkmark & \checkmark &  & $68.48_{\uparrow4.24}$ & 40.14\\
				& \checkmark & & \checkmark & $\bm{69.76}_{\uparrow5.52}$ & 61.77 \\
				\hline
			\end{tabular}
		}
	\end{center}
	\vspace{-0.4cm}
	\label{tab:ablative_tr}
\end{table}

\section{Conclusion}
In this paper, we present TransVG, a transformer-based framework for visual grounding. Instead of leveraging complex manually designed fusion modules, TransVG uses a simple stack of transformer encoders to perform the multi-modal fusion and reasoning for the visual grounding task. Extensive experiments indicate that TransVG's multi-modal transformer layers effectively perform the step-by-step fusion and reasoning, which enable TransVG to set a series of new state-of-the-art records on multiple datasets. Our TransVG serves as a new framework and exhibits huge potential for future investigation.

\paragraph{Acknowledgements} 

This work was supported in part by the National Key Research \& Development Program of China under contract 2017YFB1002202, and in part by the National Natural Science Foundation of China under Contract 61836011 \& 61632019. It was also supported by the GPU cluster built by MCC Lab of Information Science and Technology Institution, USTC.

{\small
	\bibliographystyle{ieee_fullname}
	\bibliography{egbib}
}

\end{document}